\documentclass[10pt,twocolumn,letterpaper]{article}

\usepackage{cvpr}
\usepackage{times}
\usepackage{epsfig}
\usepackage{graphicx}
\usepackage{amsmath}
\usepackage{amssymb}

\usepackage{amsmath}
\usepackage{graphicx}
\usepackage{dsfont}
\usepackage{subcaption}
\usepackage{bm}
\usepackage{arydshln}
\usepackage{color}
\usepackage{booktabs}
\usepackage{mathtools}

\usepackage[pagebackref=true,breaklinks=true,letterpaper=true,colorlinks,bookmarks=false]{hyperref}

\cvprfinalcopy % *** Uncomment this line for the final submission

\def\cvprPaperID{xxxx} % *** Enter the CVPR Paper ID here

\ifcvprfinal\fi
\begin{document}

%%%%%%%%% TITLE
\title{Few-Shot Image Recognition by Predicting Parameters from Activations}

\author{Siyuan Qiao$^{1}$~~~~Chenxi Liu$^{1}$~~~~Wei Shen$^{1,2}$~~~~Alan Yuille$^{1}$\\
Johns Hopkins University$^{1}$~~~~Shanghai University$^{2}$\\
{\tt\small \{siyuan.qiao, cxliu, alan.yuille\}@jhu.edu~~~~wei.shen@t.shu.edu.cn}
}

\maketitle
\thispagestyle{empty}

%%%%%%%%% ABSTRACT
\begin{abstract}
  In this paper, we are interested in the few-shot learning problem.
  In particular, we focus on a challenging scenario where the number of categories is large and the number of examples per novel category is very limited, \textit{e.g.} 1, 2, or 3.
  Motivated by the close relationship between the parameters and the  activations in a neural network associated with the same category, we propose a novel method that can adapt a pre-trained neural network to novel categories by directly predicting the parameters from the activations.
  Zero training is required in adaptation to novel categories, and fast inference is realized by a single forward pass.
  We evaluate our method by doing few-shot image recognition on the ImageNet dataset, which achieves the state-of-the-art classification accuracy on novel categories by a significant margin while keeping comparable performance on the large-scale categories.
  We also test our method on the MiniImageNet dataset and it strongly outperforms the previous state-of-the-art methods.
\end{abstract}

\vspace{-0.1in}
\section{Introduction}\label{sec:i}
% Object classification is fundamental for computer vision
% Recent success comes from deep learning and big data

% Image recognition is a fundamental problem for computer vision. Recent years have witnessed rapid advances in visual recognition on large-scale image datasets \cite{NIPS2012_4824,DBLP:journals/corr/SimonyanZ14a,DBLP:conf/cvpr/HeZRS16}. The underlying drives behind the great success are deep learning \cite{lecun2015deep}, large-scale image datasets, e.g.
% ImageNet~\cite{ILSVRC15}, and high performance parallel computation~\cite{NIPS2012_0598}.
% In particular, the datasets provide a large amount of data to train deep neural networks, and training can be done on accelerating infrastructures such as GPUs to be able to complete in reasonable time.
% Image recognition is a fundamental problem for computer vision.
Recent years have witnessed rapid advances in deep learning~\cite{lecun2015deep},
with a particular example being visual recognition~\cite{DBLP:conf/cvpr/HeZRS16,NIPS2012_4824,DBLP:journals/corr/SimonyanZ14a} on large-scale image datasets, \textit{e.g.}, ImageNet~\cite{ILSVRC15}.
Despite their great performances on benchmark datasets, the machines exhibit clear difference with people in the way they learn concepts.
Deep learning methods typically require huge amounts of supervised training data per concept, and the learning process could take days using specialized hardware, \textit{i.e.} GPUs.
In contrast, children are known to be able to learn novel visual concepts almost effortlessly with a few examples after they have accumulated enough past knowledge~\cite{bloom2000children}.
This phenomenon motivates computer vision research on the problem of few-shot learning, \textit{i.e.}, the task to learn novel concepts from only a few examples for each category~\cite{fei2006one,lake2015human}.

% Human's ability to learn new concepts
% Motivates a series of work with focus on few-shot learning
% In contrast to the huge amount of data and training time required by deep learning for learning new visual concepts, children are known to be able to form quick hypotheses about the meanings of the words they hear and associate them with the novel visual concepts they see almost effortlessly after they have accumulated enough past knowledge~\cite{bloom2000children}. This phenomenon motivates computer vision research on the problem of few-shot learning, i.e., the task to learn novel visual concepts from only a few examples for each category~\cite{fei2006one,lake2015human}.

Formally, in the few-shot learning problem~\cite{siamese,mao2015learning,DBLP:conf/nips/VinyalsBLKW16},
we are provided with a large-scale set
$\mathcal{D}_{\text{large}}$ with categories $\mathcal{C}_{\text{large}}$ and a few-shot set $\mathcal{D}_{\text{few}}$ with categories $\mathcal{C}_{\text{few}}$ that do not overlap with $\mathcal{C}_{\text{large}}$.
$\mathcal{D}_{\text{large}}$ has sufficient training samples for each category whereas $\mathcal{D}_{\text{few}}$ has only a few examples ($<6$ in this paper). The goal is to achieve good classification performances, either on $\mathcal{D}_{\text{few}}$ or on both $\mathcal{D}_{\text{few}}$ and $\mathcal{D}_{\text{large}}$.
We argue that a good classifier should have the following properties: (1) It achieves reasonable performance on $\mathcal{C}_{\text{few}}$. (2) Adapting to $\mathcal{C}_{\text{few}}$ does not degrade the performance on $\mathcal{C}_{\text{large}}$ significantly (if any).
(3) It is fast in inference and adapts to few-shot categories with little or zero training, \textit{i.e.}, an efficient lifelong learning system~\cite{DBLP:conf/kdd/Chen014,DBLP:conf/icml/Chen014}.

Both parametric and non-parametric methods have been proposed for the few-shot learning problem.
However, due to the limited number of samples in $\mathcal{D}_{\text{few}}$ and the imbalance between $\mathcal{D}_{\text{large}}$ and $\mathcal{D}_{\text{few}}$, parametric models usually fail to learn well from the training samples~\cite{mao2015learning}.
On the other hand, many non-parametric approaches such as nearest neighbors can adapt to the novel concepts easily without severely forgetting the original classes. But this requires careful designs of the distance metrics~\cite{atkeson1997locally}, which can be difficult and sometimes empirical.
To remedy this, some previous work instead adapts feature representation to the metrics by using siamese networks~\cite{siamese,lin2017transfer}.
As we will show later through experiments, these methods do not fully satisfy the properties mentioned above.

\begin{figure}[!htp]
    \centering
    \includegraphics[width=0.9\linewidth]{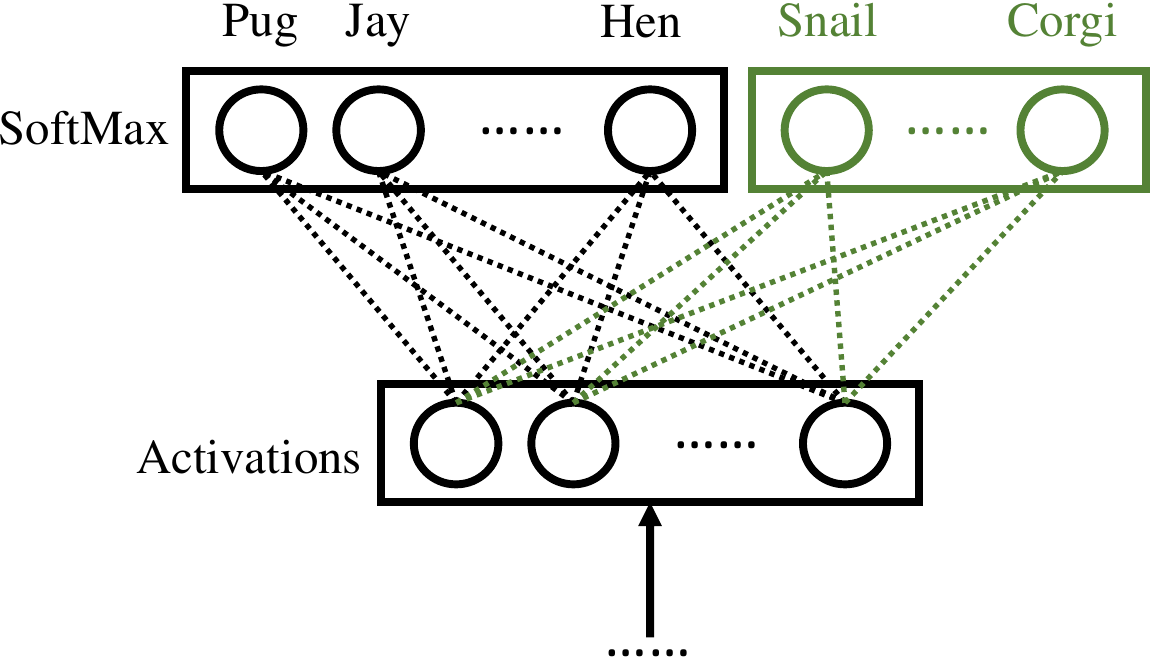}
    \caption{Illustration of pre-training on $\mathcal{D}_{\text{large}}$ (black) and few-shot novel category adaptation to $\mathcal{D}_{\text{few}}$ (green).
    The green circles are the novel categories, and the green lines represent the unknown parameters for categories in $C_{\text{few}}$.}
    \label{fig:lsfs}
\end{figure}

\begin{figure*}
    \begin{subfigure}{.499\linewidth}
        \centering
        \includegraphics[width=\linewidth]{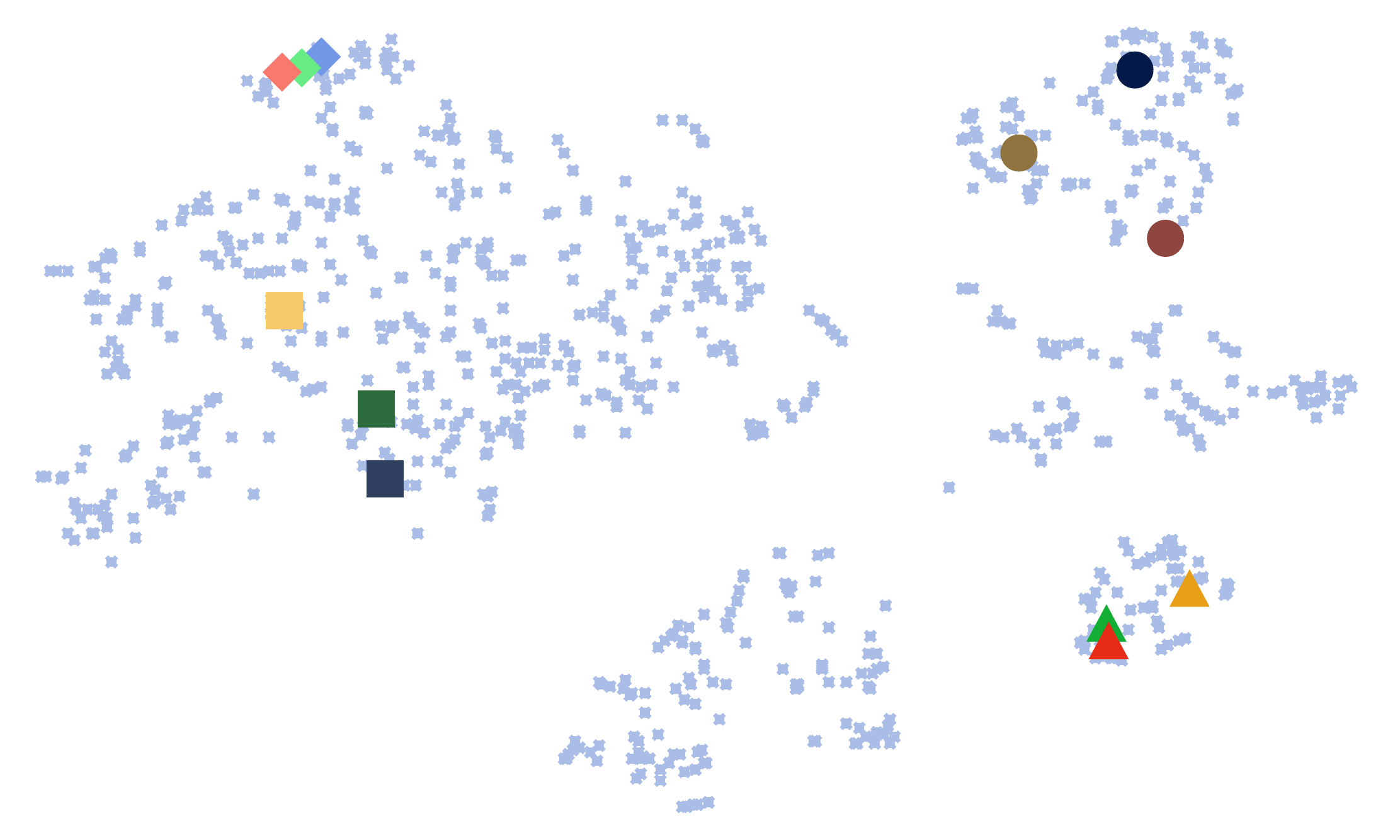}
    \end{subfigure}
    \vrule
    \begin{subfigure}{.499\linewidth}
        \centering
        \includegraphics[width=\linewidth]{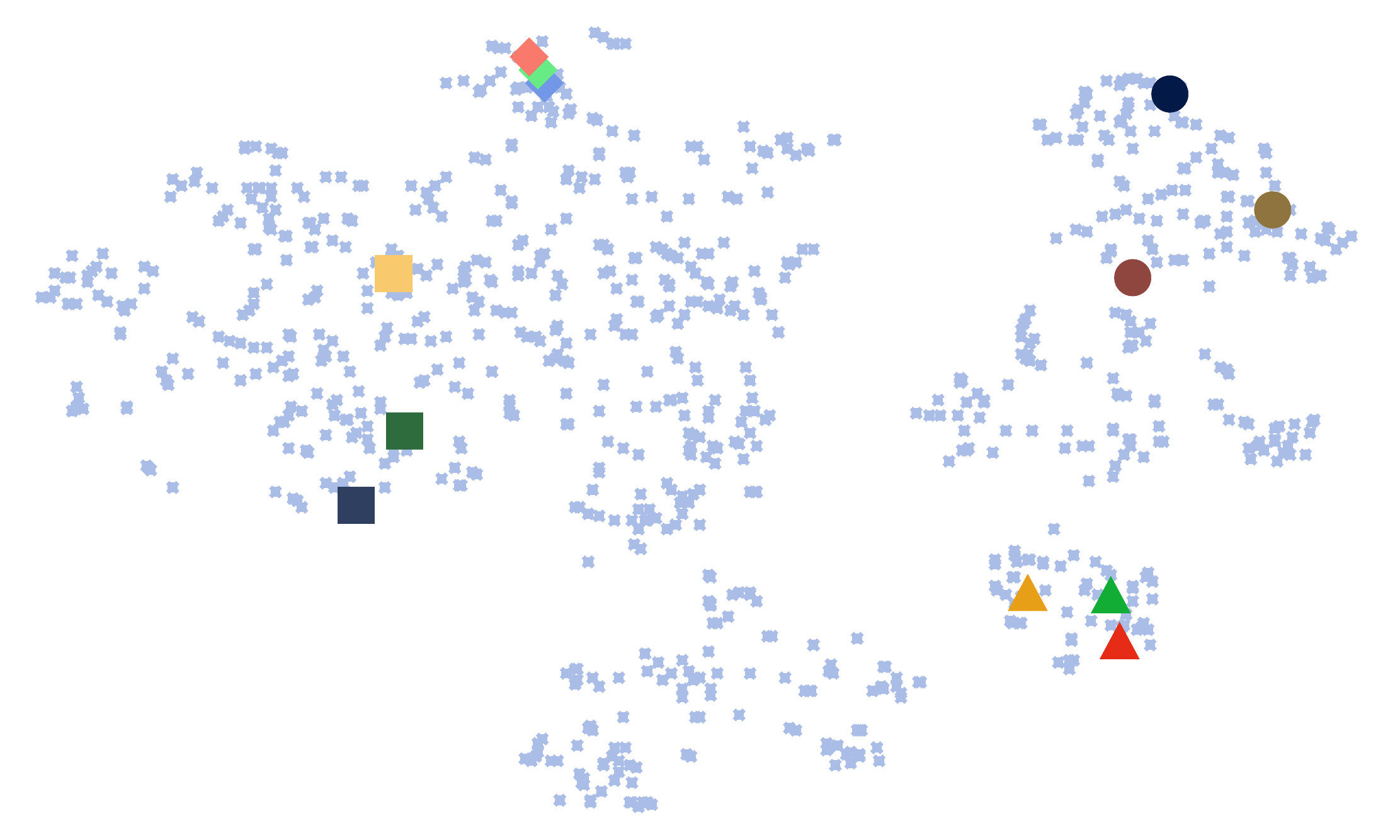}
    \end{subfigure}
    \caption{Our motivation: t-SNE~\cite{maaten2008visualizing} results on the average activations $\bar{\mathbf{a}}_y$ of each category before the fully connected layer of a $50$-layer ResNet~\cite{DBLP:conf/cvpr/HeZRS16} pre-trained on $\mathcal{D}_{\text{large}}$ from ImageNet~\cite{ILSVRC15} (left) and the parameters $\mathbf{w}_y$ of each category in the last fully connected layer (right). Each point represents a category. Highlighted points with the same color and shape correspond to the same category.
    Circles are \emph{mammals}, triangles are \emph{birds}, diamonds are \emph{buses}, and squares are \emph{home appliances.}}
    \label{fig:tsne}
\end{figure*}

\vspace{-0.1in}
% We modify learn like a child paper to make it adaptive to data scales
% we need weights, which can be obtained from features -- dL/dW = f
In this paper, we present an approach that meets the desired properties well.
% Our work is motivated by Learning like a Child~\cite{mao2015learning}, which adapts to novel concepts by updating only parameters in the deep neural network that correspond to new categories from a fixed feature embedding, e.g., convolutional layers in that network. However, since we are under few-shot constraints, we cannot learn the parameters like~\cite{mao2015learning}. Instead, we propose to learn a mapping from embedded features to parameters directly. Note that this mapping also occurs in $\mathcal{D}_{\text{large}}$, therefore has a large amount of data to train. When in the dataset $\mathcal{D}_{\text{few}}$, we use the trained mapping to regress the parameters from the embedded features of the available examples of novel categories, then extend the previous classifier by adding these parameters into it. The resulted system will operate just as a normal deep neural network except some part of the parameters are not trained by back-propagation from supervised data.
Our method starts with a pre-trained deep neural network on $\mathcal{D}_{\text{large}}$. The final classification layers (the fully connected layer and the softmax layer) are shown in Figure~\ref{fig:lsfs}.
We use $\mathbf{w}_y \in \mathbb{R}^n$ to denote the parameters for category $y$ in the fully connected layer, and use $\mathbf{a}(x) \in \mathbb{R}^n$ to denote the activations before the fully connected layer of an image $x$.
Training on $\mathcal{D}_{\text{large}}$ is standard; the real challenge is how to re-parameterize the last fully connected layer to include the novel categories under the few-shot constraints, \textit{i.e.}, for each category in $\mathcal{C}_{\text{few}}$ we have only a few examples.
Our proposed method addresses this challenge by directly predicting the parameters $\mathbf{w}_y$ (in the fully connected layer) using the activations belonging to that category, \textit{i.e.} $\mathcal{A}_y = \{ \mathbf{a}(x) | x \in \mathcal{D}_{\text{large}} \cup \mathcal{D}_{\text{few}}, Y(x) = y \}$, where $Y(\cdot)$ denotes the category of the image.

% We also observe that: (1) Nearest neighbor with cosine similarity is essentially regressing parameters from activations by an identity mapping, provided the activations are normalized.
% Intuitively, the parameters should have high correlations with the activations they correspond to while having low correlations with others.
% Let $\bm{\nu}_y \in \mathbb{R}^n$ be some statistic of $\mathcal{A}_y$, which can either be the mean of all activations in the set or a random sample from the set.
This parameter predictor stems from the tight relationship between the parameters and activations.
Intuitively in the last fully connected layer, we want $\mathbf{w}_y \cdot \mathbf{a}_y$ to be large, for all $\mathbf{a}_y \in \mathcal{A}_y$.
Let $\bar{\mathbf{a}}_y \in \mathbb{R}^n$ be the mean of the activations in $\mathcal{A}_y$.
Since it is known that the activations of images in the same category are spatially clustered together~\cite{DBLP:conf/icml/DonahueJVHZTD14}, a reasonable choice of $\mathbf{w}_y$ is to align with $\bar{\mathbf{a}}_y$ in order to maximize the inner product, and this argument holds true for all $y$.
% (which is good for few-shot).
To verify this intuition, we use t-SNE~\cite{maaten2008visualizing} to visualize the neighbor embeddings of the activation statistic $\bar{\mathbf{a}}_y$ and the parameters $\mathbf{w}_y$ for each category of a pre-trained deep neural network, as shown in Figure~\ref{fig:tsne}.
Comparing them and we observe a high similarity in both the local and the global structures.
More importantly, the semantic structures~\cite{DBLP:journals/corr/HuhAE16} are also preserved in both activations and parameters, indicating a promising generalizability to unseen categories.
% Therefore, we speculate that there exists a category-agnostic mapping from activations to parameters for a given feature embedding, and we learn this mapping using deep learning framework.

These results suggest the existence of a category-agnostic mapping from the activations to the parameters given a good feature extractor $\mathbf{a}(\cdot)$.
In our work, we parameterize this mapping with a feedforward network that is learned by back-propagation.
This mapping, once learned, is used to predict parameters for both $\mathcal{C}_{\text{few}}$ and $\mathcal{C}_{\text{large}}$.

% We evaluate our method on ILSVRC 2015~\cite{ILSVRC15}, a large-scale dataset with $1000$ categories.
% We split the categories into two sets where $\mathcal{C}_{\text{large}}$ has $900$ and $\mathcal{C}_{\text{few}}$ has the rest $100$.
% This is a setting that is considerably larger than what has been experimented in few-shot learning before.
% We compare our method with previous work and show state-of-the-art performance.

We evaluate our method on two datasets.
The first one is MiniImageNet~\cite{DBLP:conf/nips/VinyalsBLKW16}, a simplified subset of ImageNet ILSVRC 2015~\cite{ILSVRC15}, in which $C_{\text{large}}$ has $80$ categories and $C_{\text{few}}$ has $20$ categories.
Each category has $600$ images of size $84\times 84$.
This small dataset is the benchmark for natural images that the previous few-shot learning methods are evaluated on.
However, this benchmark only reports the performances on $D_{\text{few}}$, and the accuracy is evaluated under $5$-way test, \textit{i.e.}, to predict the correct category from only $5$ category candidates.
In this paper, we will take a step forward by evaluating our method on the full ILSVRC 2015~\cite{ILSVRC15}, which has $1000$ categories.
We split the categories into two sets where $\mathcal{C}_{\text{large}}$ has $900$ and $\mathcal{C}_{\text{few}}$ has the rest $100$.
The methods will be evaluated under $1000$-way test on both $D_{\text{large}}$ and $D_{\text{few}}$.
This is a setting that is considerably larger than what has been experimented in the few-shot learning before.
We compare our method with the previous work and show state-of-the-art performances.

% \vspace{-0.1in}
% \clearpage

% Organization
The rest of the paper is organized as follows: \S\ref{sec:m} defines and explains our model, \S\ref{sec:r} presents the related work, \S\ref{sec:e} shows the experimental results, and \S\ref{sec:c} concludes the paper.

\section{Model}\label{sec:m}
\begin{figure*}
    \centering
    \includegraphics[width=1.0\linewidth]{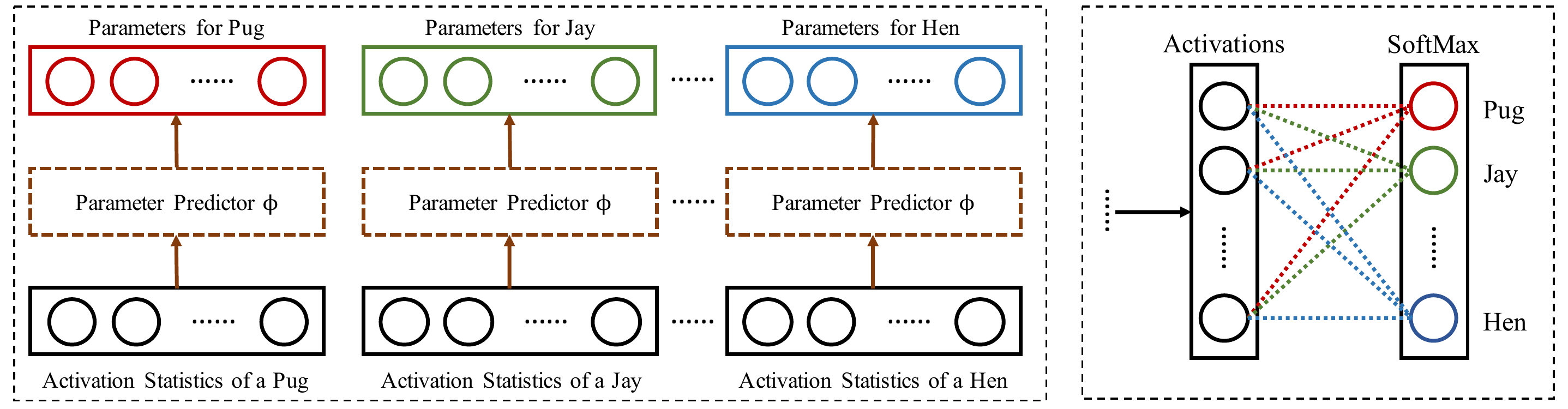}
    \caption{Building the fully connected layer by parameter prediction from activation statistics.}
    \label{fig:ppff}
\end{figure*}

The key component of our approach is the category-agnostic parameter predictor $\phi: \bar{\mathbf{a}}_y \rightarrow \mathbf{w}_y$ (Figure~\ref{fig:ppff}).
More generally, we could allow the input to $\phi$ to be a statistic representing the activations of category $y$.
Note that we use the same mapping function for all categories $y \in \mathcal{C}_{\text{large}}$, because we believe the activations and the parameters have similar local and global structure in their respective space.
Once this mapping has been learned on $\mathcal{D}_{\text{large}}$, because of this structure-preserving property, we expect it to generalize to categories in $\mathcal{C}_{\text{few}}$.
% One strategy is to take the activations computed by the feature extractor $\mathbf{a}(\cdot)$ and learn $\phi$ to minimize a loss between the pre-trained parameters $\mathbf{w}$ and $\phi(\mathbf{a})$.
% Here, we take another approach by learning $\phi$ from classification supervision.

% \vspace{-0.05in}
\subsection{Learning Parameter Predictor}
% \vspace{-0.05in}
Since our final goal is to do classification, we learn $\phi$ from the classification supervision.
Specifically, we can learn $\phi$ from $\mathcal{D}_{\text{large}}$ by minimizing the classification loss (with a regularizer $||\phi||$) defined by
\begin{equation}\label{eq:1}
    \footnotesize
    \mathcal{L}(\phi)= \sum_{(y, x) \in \mathcal{D}_{\text{large}}} \left[ -\phi\left(\bar{\mathbf{a}}_y\right)\mathbf{a}(x)+\log\sum_{\mathclap{y' \in \mathcal{C}_{\text{large}}}} e^{\phi\left(\bar{\mathbf{a}}_{y'} \right)\mathbf{a}(x)} \right]+\lambda||\phi||
\end{equation}
%\begin{equation}\label{eq:1}
%    \mathcal{L}(\phi)=\mathop\mathbb{E}_{(y,x),\mathbf{s}_1,...,\mathbf{s}_{|\mathcal{C}_{\text{large}}|}}\left[ -\phi\left(\mathbf{s}_y\right)\mathbf{a}(x)+\log\sum_i e^{\phi\left(\mathbf{s}_i\right)\mathbf{a}(x)} \right]+\lambda||\phi||_2^2
%\end{equation}

% \subsection{Learning Parameter Predictor from Sampled Activations}

Eq.~\ref{eq:1} models the parameter prediction for categories $y\in\mathcal{C}_{\text{large}}$. However, for the few-shot set $\mathcal{C}_{\text{few}}$, each category only has a few activations, whose mean value is the activation itself when each category has only one sample.
To model this few-shot setting in the large-scale training on $\mathcal{D}_{\text{large}}$, we allow both the individual activations and the mean activation to represent a category.
Concretely, let $\mathbf{s}_y \in \mathcal{A}_y \cup \bar{\mathbf{a}}_y$ be a statistic for category $y$.
Let $S_{\text{large}}$ denote a statistic set $\{\mathbf{s}_1,...,\mathbf{s}_{|\mathcal{C}_{\text{large}}|}\}$ with one for each category in $\mathcal{C}_{\text{large}} $.
We sample activations $\mathbf{s}_y$ for each category $y$ from $\mathcal{A}_y \cup \bar{\mathbf{a}}_y$ with a probability $p_{\text{mean}}$ to use $\bar{\mathbf{a}}_y$ and $1-p_{\text{mean}}$ to sample uniformly from $\mathcal{A}_y$.
Now, we learn $\phi$ to minimize the loss defined by
\begin{equation}\label{eq:mixed}
    \footnotesize
    \mathcal{L}(\phi)= \sum_{\mathclap{(y, x) \in \mathcal{D}_{\text{large}}}}~~\mathbb{E}_{S_{\text{large}}} \left[ -\phi\left(\mathbf{s}_y\right)\mathbf{a}(x)+\log\sum_{\mathclap{y' \in \mathcal{C}_{\text{large}}}} e^{\phi\left(\mathbf{s}_{y'} \right)\mathbf{a}(x)} \right]+\lambda||\phi||
\end{equation}

% \vspace{-0.05in}
\subsection{Inference}
% \vspace{-0.05in}
During inference we include $\mathcal{C}_{\text{few}}$, which calls for a statistic set for all categories $S=\{\mathbf{s}_1,...,\mathbf{s}_{|\mathcal{C}|}\}$, where $\mathcal{C} = \mathcal{C}_{\text{large}} \cup \mathcal{C}_{\text{few}}$.
Each statistic set $S$ can generate a set of parameters $\{ \phi(\mathbf{s}_1),...,\phi(\mathbf{s}_{|\mathcal{C}|}) \}$ that can be used for building a classifier on $\mathcal{C}$.
Since we have more than one possible set $S$ from the dataset $\mathcal{D} = \mathcal{D}_{\text{large}} \cup \mathcal{D}_{\text{few}}$, we can do classification based on all the possible $S$.
% Note that here in inference we include the few-shot categories as well, instead of using only $\mathcal{D}_{\text{large}}$ as in training.
Formally, we compute the probability of $x$ being in category $y$ by
\begin{equation}\label{eq:2}
    \footnotesize
    P(y|x)=e^{\mathop\mathbb{E}_{S}\left[ \phi(\mathbf{s}_y) \mathbf{a}(x) \right]}/\left(\sum_{y' \in \mathcal{C}} e^{\mathop\mathbb{E}_{S}\left[ \phi(\mathbf{s}_{y'}) \mathbf{a}(x) \right]}\right)
\end{equation}

% Note that here we do not restrict the dataset $\mathcal{D}$ to be our training dataset $\mathcal{D}_{\text{large}}$. Setting $\mathcal{D}=\mathcal{D}_{\text{large}}\cup\mathcal{D}_{\text{few}}$ automatically adapts the neural network towards the novel categories without forgetting the known ones.
However, classifying images with the above equation is time-consuming since it computes the expectations over the entire space of $S$ which is exponentially large.
We show in the following that if we assume $\phi$ to be a linear mapping, then this expectation can be computed efficiently.
% To reduce the computation time to the level of a normal neural network, we first assume $\phi$ is linear, then generalize to non-linear regression.

% \subsection{Efficient Inference and Metric Learning}\label{sec:ei}
In the linear case $\phi$ is a matrix $\Phi$.
The predicted parameter for category $y$ is
\begin{equation}\label{eq:3}
    \footnotesize
    \hat{\mathbf{w}}_y = \Phi \cdot \mathbf{s}_y
\end{equation}
The inner product of $x$ before the softmax function for category $y$ is
\begin{equation}
    \footnotesize
    h (\mathbf{s}_y, \mathbf{a}(x)) = \hat{\mathbf{w}}_y \cdot \mathbf{a}(x) = \Phi \cdot \mathbf{s}_y \cdot \mathbf{a}(x)
\end{equation}
If $\mathbf{a}(x)$ and $\mathbf{s}_y$ are normalized, then by setting $\Phi$ as the identity matrix, $h (\mathbf{s}_y, \mathbf{a}(x))$ is equivalent to the cosine similarity between $\mathbf{s}_y$ and $\mathbf{a}(x)$.
Essentially, by learning $\Phi$, we are learning a more general similarity metric on the activations $\mathbf{a}(x)$ by capturing correlations between different dimensions of the activations.
We will show more comparisons between the learned $\Phi$ and identity matrix in \S\ref{sec:cai}.
% Compared with cosine similarity, however, our model discovers more correlation within $\mathbf{a}(x)$.
%Here, we learn $\phi_{ij}$ from $\mathcal{D}_{\text{large}}$ by minimizing the loss
%\begin{equation}\label{eq:l}
%    \mathcal{L}(\phi)=\mathop\mathbb{E}_{(y,x),r_1,...,r_{|\mathcal{C}_{\text{large}}|}}\left[ -\sum_{ij}\phi_{ij}a_i(r_y)a_j(x) +\log\sum_k e^{\sum_{ij}\phi_{ij}a_i(r_k)a_j(x)} \right]+\lambda ||\phi||_2^2
%\end{equation}

% \begin{align}\label{eq:6}
%     \begin{split}
%     P(y|x)&=e^{\mathop\mathbb{E}_{R}\left[ \phi(\mathbf{a}(r_y)) \mathbf{a}(x) \right]}/\left(\sum_i e^{\mathop\mathbb{E}_{R}\left[ \phi(\mathbf{a}(r_i)) \mathbf{a}(x) \right]}\right)\\
%     &=e^{\mathbf{a}(x) \phi(\mathop\mathbb{E}_{R}\left[ \mathbf{a}(r_y)  \right])}/\left(\sum_i e^{\mathbf{a}(x) \phi(\mathop\mathbb{E}_{R} \left[ \mathbf{a}(r_i) \right])}\right)
%     \end{split}
%     % P(y|x)=e^{\mathbf{a}(x) \phi(\mathop\mathbb{E}_{R}\left[ \mathbf{a}(r_y)  \right])}/\left(\sum_i e^{\mathbf{a}(x) \phi(\mathop\mathbb{E}_{R} \left[ \mathbf{a}(r_i) \right])}\right)
% \end{align}

% -- chenxi
% \begin{align}
%     P(y|x)
%     &=e^{\mathbf{a}(x) \cdot \phi(\mathop\mathbb{E}_{S}\left[\mathbf{s}_y\right])  }/\left(\sum_{y' \in \mathcal{C}} e^{\mathbf{a}(x) \cdot \phi( \mathop\mathbb{E}_{S}\left[ \mathbf{s}_{y'}  \right] ) }\right) \label{eq:6} \\
%     &=e^{\mathbf{a}(x) \cdot \Phi \cdot \mathop\mathbb{E}_{S}[ \mathbf{s}_y  ]}/\left(\sum_{y' \in \mathcal{C}} e^{\mathbf{a}(x) \cdot \Phi \cdot \mathop\mathbb{E}_{S} \left[ \mathbf{s}_{y'} \right]}\right) \label{eq:7}
% \end{align}
Because of the linearity of $\phi$, the probability of $x$ being in category $y$ simplifies to
\begin{equation}\label{eq:7}
    \footnotesize
    \begin{aligned}
    P(y|x)
    = & ~ e^{\mathbf{a}(x) \cdot \phi(\mathop\mathbb{E}_{S}\left[\mathbf{s}_y\right])  }/\left(\sum_{y' \in \mathcal{C}} e^{\mathbf{a}(x) \cdot \phi( \mathop\mathbb{E}_{S}\left[ \mathbf{s}_{y'}  \right] ) }\right) \\
    = & ~ e^{\mathbf{a}(x) \cdot \Phi \cdot \mathop\mathbb{E}_{S}[ \mathbf{s}_y  ]}/\left(\sum_{y' \in \mathcal{C}} e^{\mathbf{a}(x) \cdot \Phi \cdot \mathop\mathbb{E}_{S} \left[ \mathbf{s}_{y'} \right]}\right)
    \end{aligned}
\end{equation}
Now $\mathbb{E}_S [\mathbf{s}_y]$ can be pre-computed which is efficient.
Adapting to novel categories only requires updating the corresponding $\mathbb{E}_S [\mathbf{s}_y]$.
Although it is ideal to keep the linearity of $\phi$ to reduce the amount of computation, introducing non-linearity could potentially improve the performance.
To keep the efficiency, we still push in the expectation and approximate Eq.~\ref{eq:2} as in Eq.~\ref{eq:7}.

\begin{figure*}
  \begin{subfigure}[b]{.459\linewidth}
      \centering
      \includegraphics[width=\linewidth]{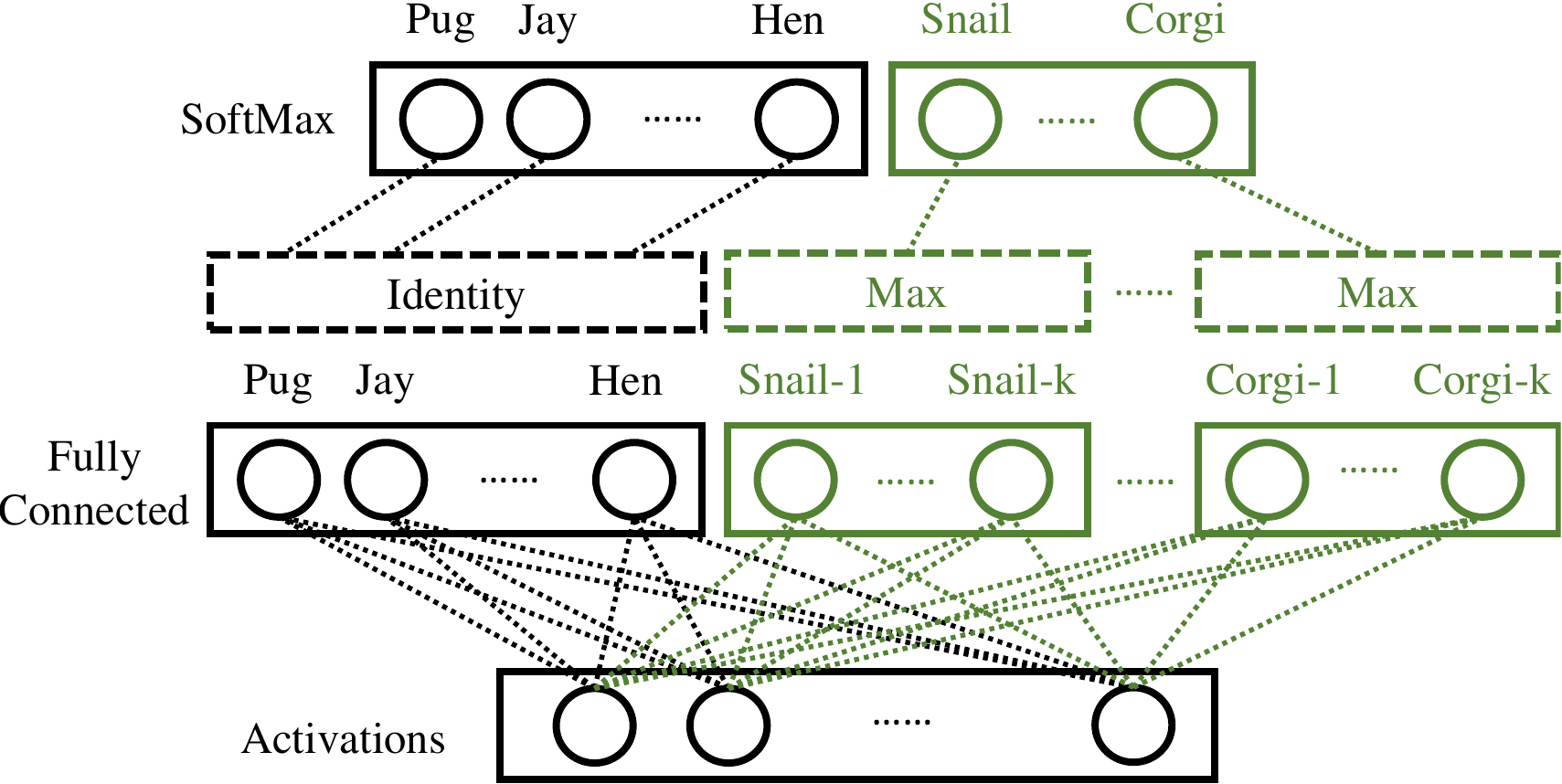}
      \caption{}\label{fig:addnew}
  \end{subfigure}
  \hfill
  \begin{subfigure}[b]{.499\linewidth}
      \centering
      \includegraphics[width=\linewidth]{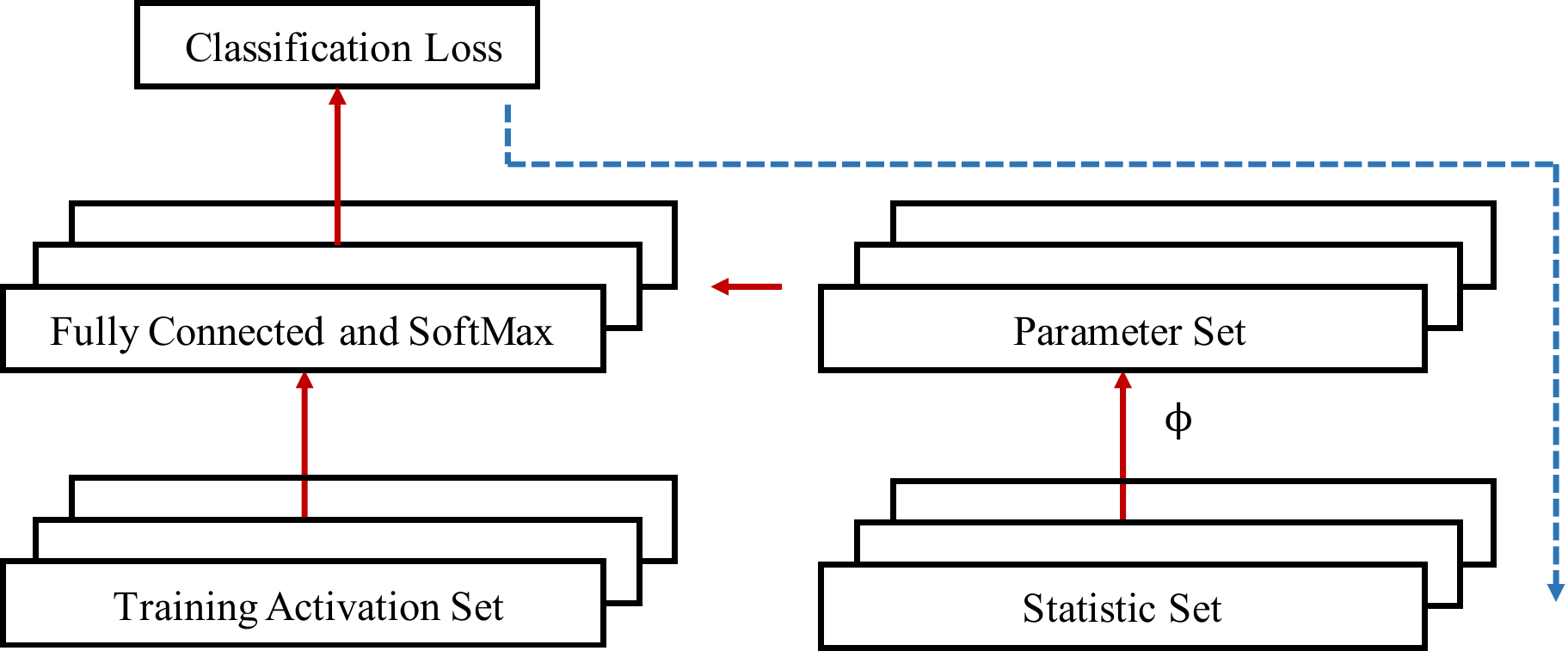}
      \caption{}\label{fig:train}
  \end{subfigure}
    \caption{Illustration of the novel category adaption (a) and the training strategies for parameter predictor $\phi$ (b).
    (b): red and solid arrows show the feedforward data flow, while blue and dashed arrow shows the backward gradient flow.}
    \label{fig:addnewandtraining}
\end{figure*}
When adding categories $y\in\mathcal{C}_{\text{few}}$, the estimate of $\mathbb{E}_S [\mathbf{s}_y]$ may not be reliable since the number of samples is small.
Besides, Eq.~\ref{eq:mixed} models the sampling from one-shot and mean activations.
Therefore, we take a mixed strategy for parameter prediction, \textit{i.e.}, we use $\mathop\mathbb{E}_S[\mathbf{s}_y]$ to predict parameters for category $y\in\mathcal{C}_{\text{large}}$,
but for $\mathcal{C}_{\text{few}}$ we treat each sample as a newly added category, as shown in Figure~\ref{fig:addnew}.
% The inputs to the SoftMax layer for the novel categories are mixture models of the corresponding reference activations.
For each novel category in $C_{\text{few}}$, we compute the maximal response of the activation of the test image to the parameter set predicted from each activation in the statistic set of the corresponding novel category in $C_{\text{few}}$.
We use them as the inputs to the SoftMax layer to compute the probabilities.

% \vspace{-0.05in}
\subsection{Training Strategy}\label{sec:ts}
The objective of training is to find $\phi$ that minimizes Eq.~\ref{eq:mixed}. There are many methods to do this. We approach this by using stochastic gradient decent with weight decay and momentum.
Figure~\ref{fig:train} demonstrates the training strategy of the parameter predictor $\phi$.
We train $\phi$ on $\mathcal{D}_{\text{large}}$ with categories $\mathcal{C}_{\text{large}}$. For each batch of the training data, we sample $|\mathcal{C}_{\text{large}}|$ statistics $\mathbf{s}_y$ from $\mathcal{A}_y \cup \bar{\mathbf{a}}_y$ to build a statistic set $S$ with one for each category $y$ in $\mathcal{C}_{\text{large}}$.
Next, we sample a training activation set $T$ from $\mathcal{D}_{\text{large}}$ with one for each category in $\mathcal{C}_{\text{large}}$.
In total, we sample $2|\mathcal{C}_{\text{large}}|$ activations.
The activations in the statistic sets are fed to $\phi$ to generate parameters for the fully connected layer. With the predicted parameters for each category in $\mathcal{C}_{\text{large}}$, the training activation set then is used to evaluate their effectiveness by classifying the training activations. At last, we compute the classification loss with respect to the ground truth, based on which we calculate the gradients and back-propagate them in the path shown in Figure~\ref{fig:train}. After the gradient flow passes through $\phi$, we update $\phi$ according to the gradients.
% and one iteration of training is done.

% CHANGE 'TEST ACTIVATION SET' TO 'TRAINING ACTIVATION SET'. HAVING A 'TEST ACTIVATION SET' IN THE TRAINING STAGE IS CONFUSING. ALSO FIX THIS IN THE FOLLOWING SECTION 2.4.

% \vspace{-0.05in}
\subsection{Implementation Details}\label{sec:id}
% \vspace{-0.05in}
\paragraph{Full ImageNet Dataset}
Our major experiments are conducted on ILSVRC 2015~\cite{ILSVRC15}. ILSVRC 2015 is a large-scale image dataset with $1000$ categories, each of which has about $1300$ images for training, and $50$ images for validation. For the purpose of studying both the large-scale and the few-shot settings at the same time, ILSVRC 2015 is split to two sets by the categories.
The training data from $900$ categories are collected into $\mathcal{D}_{\text{large}}$, while the rest $100$ categories are gathered as set $\mathcal{D}_{\text{few}}$.
% According to the alphabetical order, the top $900$ categories are collected into $\mathcal{D}_{\text{large}}$, while the rest $100$ categories are gathered as set $\mathcal{D}_{\text{few}}$.

% MAYBE BETTER NOT TO MENTION HOW THESE 100 CATEGORIES ARE CHOSEN. SINCE YOU DID NOT CHOOSE RANDOMLY, THIS MAY CAUSE PROBLEM.

We first train a $50$-layer ResNet~\cite{DBLP:conf/cvpr/HeZRS16} on $\mathcal{D}_{\text{large}}$. We use the outputs of the global average pooling layer as the activation $\mathbf{a}(x)$ of an image $x$. For efficiency, we compute the activation $\mathbf{a}(x)$ for each image $x$ before the experiments as well as the mean activations $\bar{\mathbf{a}}_y$.
Following the training strategy shown in \S\ref{sec:ts}, for each batch, we sample $900$ activations as the statistic set and $900$ activations as the training activation set. We compute the parameters using the statistic set, and copy the parameters into the fully connected layer.
Then, we feed the training activations into the fully connected layer, calculate the loss and back-propagate the gradients. Next, we redirect the gradient flow into $\phi$. Finally, we update $\phi$ using stochastic gradient descent. The learning rate is set to $0.001$. The weight decay is set to $0.0005$ and the momentum is set to $0.9$. We train $\phi$ on $\mathcal{D}_{\text{large}}$ for $300$ epochs, each of which has $250$ batches. $p_{\text{mean}}$ is set to $0.9$.

For the parameter predictor, we implement three different $\phi$: $\phi^1$, $\phi^2$ and $\phi^{2*}$. $\phi^1$ is a one-layer fully connected model. $\phi^2$ is defined as a sequential network with two fully connected layers in which each maps from $2048$ dimensional features to $2048$ dimensional features and the first one is followed by a ReLU non-linearity layer~\cite{DBLP:conf/icml/NairH10}. The final outputs are normalized to unity in order to speed up training and ensure generalizability. By introducing non-linearity, we observe slight improvements on the accuracies for both $\mathcal{C}_{\text{large}}$ and $\mathcal{C}_{\text{few}}$.
To demonstrate the effect of minimizing Eq.~\ref{eq:mixed} instead of Eq.~\ref{eq:1}, we train another $\phi^{2*}$ which has the same architecture with $\phi^{2}$ but minimizes Eq.~\ref{eq:1}. As we will show later through experiments, $\phi^{2*}$ has strong bias towards $\mathcal{C}_{\text{large}}$.

% \begin{figure}
%         \centering
%         \includegraphics[width=1.0\linewidth]{img/train.pdf}
%     \caption{Illustration of the training strategies for $\phi$. Red and solid arrows show the feedforward data flow. Blue and dashed arrow shows the backward gradient flow.}
%     \label{fig:train}
% \end{figure}

\paragraph{MiniImageNet Dataset}
For comparison purposes, we also test our method on MiniImageNet dataset~\cite{DBLP:conf/nips/VinyalsBLKW16}, a simplified subset of ILSVRC 2015.
This dataset has $80$ categories for $D_{\text{large}}$ and $20$ categories for $D_{\text{few}}$.
Each category has $600$ images.
Each image is of size $84\times 84$.
For the fairness of comparisons, we train two convolutional neural networks to get the activation functions $\mathbf{a}(\cdot)$.
The first one is the same as that of Matching Network~\cite{DBLP:conf/nips/VinyalsBLKW16}, and the second one is a wide residual network~\cite{Zagoruyko2016WRN}.
We train the wide residual network WRN-$28$-$10$~\cite{Zagoruyko2016WRN} on $D_\text{large}$, following its configuration for CIFAR-100 dataset~\cite{cifar}.
There are some minor modifications to the network architecture as the input size is different.
To follow the architecture, the input size is set to $80\times 80$.
The images will be rescaled to this size before training and evaluation.
There will be $3$ times of downsampling rather than $2$ times as for CIFAR dataset.
The training process follows WRN-$28$-$10$~\cite{Zagoruyko2016WRN}.
We also use the output of the global average pooling layer as the activation $\mathbf{a}(x)$ of an image $x$.
For the parameter predictor $\phi$, we train it by following the settings of $\phi^2$ for the full ImageNet dataset except that now the dimension corresponds to the output of the activations of the convolutional neural networks.
The two architectures will be detailed in the appendix.

% When adding novel categories, we take a mixed strategy for parameter prediction, i.e., we use $\mathop\mathbb{E}_S[\mathbf{s}_y]$ to predict parameters for category $y\in\mathcal{C}_{\text{large}}$,
% but for $\mathcal{C}_{\text{few}}$ we treat each sample as one category. This is because Eq.~\ref{eq:mixed} only models mixing one-shot and mean activations.
% By doing this, if for each category in $\mathcal{C}_{\text{few}}$ we have $m$ samples, then the classification layer will have $|\mathcal{C}_{\text{large}}|+m\cdot|\mathcal{C}_{\text{few}}|$ outputs. The probability of one category of $\mathcal{C}_{\text{few}}$ will be the maximization over the corresponding $m$ outputs.

% \vspace{-0.05in}
\section{Related Work}\label{sec:r}
% \vspace{-0.05in}
\subsection{Large-Scale Image Recognition}\label{sec:lsir}
% \vspace{-0.05in}
We have witnessed an evolution of image datasets over the last few decades. The sizes of the early datasets are relatively small. Each dataset usually collects images on the order of tens of thousands. Representative datasets include Caltech-101~\cite{fei2006one}, Caltech-256~\cite{caltech256}, Pascal VOC~\cite{pascal-voc-2007}, and CIFAR-10/100~\cite{cifar}.
Nowadays, large-scale datasets are available with millions of detailed image annotations, \textit{e.g.} ImageNet~\cite{ILSVRC15} and MS COCO~\cite{DBLP:journals/corr/LinMBHPRDZ14}.
With datasets of this scale, machine learning methods that have large capacity start to prosper, and the most successful ones are convolutional neural network based~\cite{DBLP:conf/cvpr/HeZRS16,DBLP:journals/corr/HuangLW16a,NIPS2012_4824,DBLP:journals/corr/SimonyanZ14a,sort}.

% \vspace{-0.05in}
\subsection{Few-Shot Image Recognition}
% \vspace{-0.05in}
Unlike large-scale image recognition, the research on few-shot learning has received limited attention from the community due to its inherent difficulty, thus is still at an early stage of development. As an early attempt, Fei-Fei \textit{et al.} proposed a variational Bayesian framework for one-shot image classification~\cite{fei2006one}. A method called Hierarchical Bayesian Program Learning~\cite{NIPS2013_5128} was later proposed to specifically approach the one-shot problem on character recognition by a generative model. On the same character recognition task, Koch \textit{et al.} developed a siamese convolutional network~\cite{siamese} to learn the representation from the dataset and modeled the few-shot learning as a verification task. Later, Matching Network~\cite{DBLP:conf/nips/VinyalsBLKW16} was proposed to approach the few-shot learning task by modeling the problem as a $k$-way $m$-shot image retrieval problem using attention and memory models.
Following this work, Ravi and Larochelle proposed a LSTM-based meta-learner optimizer~\cite{ravi2017optimization}, and Chelsea \textit{et al.} proposed a model-agnostic meta learning method~\cite{pmlr-v70-finn17a}.
Although they show state-of-the-art performances on their few-shot learning tasks, they are not flexible for both large-scale and few-shot learning since $k$ and $m$ are fixed in their architectures.
We will compare ours with these methods on their tasks for fair comparisons.

% \vspace{-0.05in}
\subsection{Unified Approach}
% \vspace{-0.05in}
Learning a metric then using nearest neighbor \cite{siamese,lin2017transfer,DBLP:conf/iccv/WangG15} is applicable but not necessarily optimal to the unified problem of large-scale and few-shot learning since it is possible to train a better model on the large-scale part of the dataset using the methods in \S\ref{sec:lsir}. Mao \textit{et al.} proposed a method called Learning like a Child~\cite{mao2015learning} specifically for fast novel visual concept learning using hundreds of examples per category while keeping the original performance. However, this method is less effective when the training examples are extremely insufficient, \textit{e.g.} $<6$ in this paper.

% MAYBE BETTER TO INDICATE EXACTLY HOW MANY EXAMPLES THEY USE (100) AND WE USE (< 4)

% \vspace{-0.05in}
\section{Results}\label{sec:e}
% \vspace{-0.05in}
\subsection{Full ImageNet Classification}
In this section we describe our experiments and compare our approach with other strong baseline methods.  As stated in \S\ref{sec:i}, there are three aspects to consider in evaluating a method: (1) its performance on the few-shot set $\mathcal{D}_{\text{few}}$, (2) its performance on the large-scale set $\mathcal{D}_{\text{large}}$, and (3) its computation overhead of adding novel categories and the complexity of image inference. In the following paragraphs, we will cover the settings of the baseline methods, compare the performances on the large-scale and the few-shot sets, and discuss their efficiencies.

% \vspace{-0.05in}
\paragraph{Baseline Methods}
% \vspace{-0.05in}
The baseline methods must be applicable to both large-scale and few-shot learning settings. We compare our method with a fine-tuned $50$-layer ResNet~\cite{DBLP:conf/cvpr/HeZRS16}, Learning like a Child~\cite{mao2015learning} with a pre-trained $50$-layer ResNet as the starting network, Siamese-Triplet Network~\cite{siamese,lin2017transfer} using three $50$-layer ResNets with shared parameters, and the nearest neighbor using the pre-trained $50$-layer ResNet convolutional features. We will elaborate individually on how to train and use them.

\begin{table*}
    \small
    \centering
    \setlength{\tabcolsep}{1.0em}
    \begin{tabular}{lrccll:rr}
        \toprule
        Method & $\mathcal{D}_{\text{large}}$ & $\mathcal{D}_{\text{few}}$ & FT & Top-1 $\mathcal{C}_{\text{large}}$ & Top-5 $\mathcal{C}_{\text{large}}$ & Top-1 $\mathcal{C}_{\text{few}}$ & Top-5 $\mathcal{C}_{\text{few}}$\\
        \midrule
        NN + Cosine  & 100\% & 1 & N & 71.54\% & 91.20\% & 1.72\% & 5.86\% \\
        NN + Cosine  &  10\% & 1 & N & 67.68\% & 88.90\% & 4.42\% & 13.36\% \\
        NN + Cosine  &   1\% & 1 & N & 61.11\% & 85.11\% & 10.42\% & 25.88\% \\
        Triplet Network~\cite{siamese,lin2017transfer}  & 100\% & 1 & N & 70.47\% & 90.61\% & 1.26\% & 4.94\% \\
        Triplet Network~\cite{siamese,lin2017transfer}   &  10\% & 1 & N & 66.64\% & 88.42\% & 3.48\% & 11.40\% \\
        Triplet Network~\cite{siamese,lin2017transfer}   &   1\% & 1 & N & 60.09\% & 84.83\% & 8.84\% & 22.24\% \\
        Fine-Tuned ResNet~\cite{DBLP:conf/cvpr/HeZRS16} & 100\% & 1 & Y & 76.28\% & 93.17\% & 2.82\% & 13.30\% \\
        Learning like a Child~\cite{mao2015learning} & 100\% & 1 & Y & 76.71\% & 93.24\% & 2.90\% & 17.14\% \\
        \hdashline
        Ours-$\phi^1$         & 100\% & 1 & N & 72.56\% & 91.12\% & {\bf\color{blue}19.88\%} & {\bf\color{blue}43.20\%} \\
        Ours-$\phi^2$         & 100\% & 1 & N & 74.17\% & 91.79\% & {\bf\color{red}21.58\%} & {\bf\color{red}45.82\%} \\
        Ours-$\phi^{2*}$      & 100\% & 1 & N & 75.63\% & 92.92\% & 14.32\% & 33.84\% \\
        \midrule
        NN + Cosine  & 100\% & 2 & N & 71.54\% & 91.20\% & 3.34\% & 9.88\% \\
        NN + Cosine  &  10\% & 2 & N & 67.66\% & 88.89\% & 7.60\% & 19.94\% \\
        NN + Cosine  &   1\% & 2 & N & 61.04\% & 85.04\% & 15.14\% & 35.70\% \\
        Triplet Network~\cite{siamese,lin2017transfer}  & 100\% & 2 & N & 70.47\% & 90.61\% & 2.34\% & 8.30\% \\
        Triplet Network~\cite{siamese,lin2017transfer}  &  10\% & 2 & N & 66.63\% & 88.41\% & 6.10\% & 17.46\% \\
        Triplet Network~\cite{siamese,lin2017transfer}  &   1\% & 2 & N & 60.02\% & 84.74\% & 13.42\% & 32.38\% \\
        Fine-Tuned ResNet~\cite{DBLP:conf/cvpr/HeZRS16} & 100\% & 2 & Y & 76.27\% & 93.13\% & 10.32\% & 30.34\% \\
        Learning like a Child~\cite{mao2015learning} & 100\% & 2 & Y & 76.68\% & 93.17\% & 11.54\% & 37.68\% \\
        \hdashline
        Ours-$\phi^1$         & 100\% & 2 & N & 71.94\% & 90.62\% & {\bf\color{blue}25.54\%} & {\bf\color{blue}52.98\%} \\
        Ours-$\phi^2$         & 100\% & 2 & N & 73.43\% & 91.13\% & {\bf\color{red}27.44\%} & {\bf\color{red}55.86\%} \\
        Ours-$\phi^{2*}$      & 100\% & 2 & N & 75.44\% & 92.74\% & 18.70\% & 43.92\% \\
        \midrule
        NN + Cosine  & 100\% & 3 & N & 71.54\% & 91.20\% & 4.58\% & 12.72\% \\
        NN + Cosine  &  10\% & 3 & N & 67.65\% & 88.88\% & 9.86\% & 24.96\% \\
        NN + Cosine  &   1\% & 3 & N & 60.97\% & 84.95\% & 18.68\% & 42.16\% \\
        Triplet Network~\cite{siamese,lin2017transfer}  & 100\% & 3 & N & 70.47\% & 90.61\% & 3.22\% & 11.48\% \\
        Triplet Network~\cite{siamese,lin2017transfer}  &  10\% & 3 & N & 66.62\% & 88.40\% & 8.52\% & 22.52\% \\
        Triplet Network~\cite{siamese,lin2017transfer}  &   1\% & 3 & N & 59.97\% & 84.66\% & 17.08\% & 38.06\% \\
        Fine-Tuned ResNet~\cite{DBLP:conf/cvpr/HeZRS16} & 100\% & 3 & Y & 76.25\% & 93.07\% & 16.76\% & 39.92\% \\
        Learning like a Child~\cite{mao2015learning} & 100\% & 3 & Y & 76.55\% & 93.00\% & 18.56\% & 50.70\% \\
        \hdashline
        Ours-$\phi^1$         & 100\% & 3 & N & 71.56\% & 90.21\% & {\bf\color{blue}28.72\%} & {\bf\color{blue}58.50\%} \\
        Ours-$\phi^2$         & 100\% & 3 & N & 72.98\% & 90.59\% & {\bf\color{red}31.20\%} & {\bf\color{red}61.44\%} \\
        Ours-$\phi^{2*}$      & 100\% & 3 & N & 75.34\% & 92.60\% & 22.32\% & 49.76\% \\
        \bottomrule
    \end{tabular}
    \caption{Comparing 1000-way accuracies with feature extractor $\mathbf{a}(\cdot)$ pre-trained on $\mathcal{D}_{\text{large}}$. For different $\mathcal{D}_{\text{few}}$ settings, red: the best few-shot accuracy, and blue: the second best.}
    \label{tab:real1}
\end{table*}

% IS IT BETTER TO MAKE PT IN R\_LARGE, PT SUPERSCRIPT?

As mentioned in \S\ref{sec:id}, we first train a $900$-category classifier on $\mathcal{D}_{\text{large}}$.
We will build other baseline methods using this classifier as the staring point. For convenience, we denote this classifier as \smash{$\mathcal{R}_\text{large}^{\text{pt}}$}, where pt stands for ``pre-trained''.
Next, we add the novel categories $\mathcal{C}_{\text{few}}$ to each method. For the $50$-layer ResNet, we fine tune \smash{$\mathcal{R}_\text{large}^{\text{pt}}$} with the newly added images by extending the fully connected layer to generate $1000$ classification outputs. Note that we will limit the number of training samples of $\mathcal{C}_{\text{few}}$ for the few-shot setting.
% All the parameters are updated after fine-tuning.
For Learning like a Child, however, we fix the layers before the global average pooling layer, extend the fully connected layer to include $1000$ classes, and only update the parameters for $\mathcal{C}_{\text{few}}$ in the last classification layer.
Since we have the full access to $\mathcal{D}_{\text{large}}$, we do not need Baseline Probability
Fixation~\cite{mao2015learning}. The nearest neighbor with cosine distance can be directly used for both tasks given the pre-trained deep features.

The other method we compare is Siamese-Triplet Network ~\cite{siamese,lin2017transfer}. Siamese network is proposed to approach the few-shot learning problem on Omniglot dataset~\cite{DBLP:conf/cogsci/LakeSGT11}.
In our experiments, we find that its variant Triplet Network~\cite{lin2017transfer,DBLP:conf/iccv/WangG15} is more effective since it learns feature representation from relative distances between positive and negative pairs instead of directly doing binary classification from the feature distance. Therefore, we use the Triplet Network from~\cite{lin2017transfer} on the few-shot learning problem, and upgrade its body net to the pre-trained \smash{$\mathcal{R}_\text{large}^{\text{pt}}$}. We use cosine distance as the distance metric and fine-tune the Triplet Network. For inference, we use nearest neighbor with cosine distance.
We use some techniques to improve the speed, which will be discussed later in the efficiency analysis.

% \vspace{-0.05in}
\paragraph{Few-Shot Accuracies}
% \vspace{-0.05in}
We first investigate the few-shot learning setting where we only have several training examples for $\mathcal{C}_{\text{few}}$. Specifically, we study the performances of different methods when $\mathcal{D}_{\text{few}}$ has for each category 1, 2, and 3 samples.
It is worth noting that our task is much harder than the previously studied few-shot learning: we are evaluating the top predictions out of $1000$ candidate categories, \textit{i.e.}, $1000$-way accuracies while previous work is mostly interested in $5$-way or $20$-way accuracies~\cite{pmlr-v70-finn17a,siamese,lin2017transfer,ravi2017optimization,DBLP:conf/nips/VinyalsBLKW16}.

With the pre-trained \smash{$\mathcal{R}_\text{large}^{\text{pt}}$}, the training samples in $\mathcal{D}_{\text{few}}$ are like invaders to the activation space for  $\mathcal{C}_{\text{large}}$.
Intuitively, there will be a trade-off between the performances on $\mathcal{C}_{\text{large}}$ and $\mathcal{C}_{\text{few}}$. This is true especially for non-parametric methods. Table~\ref{tab:real1} shows the performances on the validation set of ILSVRC 2015~\cite{ILSVRC15}. The second column is the percentage of data of $\mathcal{D}_{\text{large}}$ in use, and the third column is the number of samples used for each category in $\mathcal{D}_{\text{few}}$.
Note that fine-tuned ResNet~\cite{DBLP:conf/cvpr/HeZRS16} and Learning like a Child~\cite{mao2015learning} require fine-tuning while others do not.

Triplet Network is designed to do few-shot image inference by learning feature representations that adapt to the chosen distance metric. It has better performance on $\mathcal{C}_{\text{few}}$ compared with the fine-tuned ResNet and Learning like a Child when the percentage of $\mathcal{D}_{\text{large}}$ in use is low. However, its accuracies on $\mathcal{C}_{\text{large}}$ are sacrificed a lot in order to favor few-shot accuracies. We also note that if full category  supervision is provided, the activations of training a classifier do better than that of training a Triplet Network. We speculate that this is due to the less supervision of training a Triplet Network which uses losses based on fixed distance preferences.
Fine-tuning and Learning like a Child are training based, thus are able to keep the high accuracies on $\mathcal{D}_{\text{large}}$, but perform badly on $\mathcal{D}_{\text{few}}$ which does not have sufficient data for training.
Compared with them, our method shows state-of-the-art accuracies on $\mathcal{C}_{\text{few}}$ without compromising too much the performances on $\mathcal{C}_{\text{large}}$.

Table~\ref{tab:real1} also compares $\phi^2$ and $\phi^{2*}$, which are trained to minimize Eq.~\ref{eq:mixed} and Eq.~\ref{eq:1}, respectively. Since during training $\phi^{2*}$ only mean activations are sampled, it shows a bias towards $\mathcal{C}_{\text{large}}$. However, it still outperforms other baseline methods on $\mathcal{C}_{\text{few}}$.
In short, modeling using Eq.~\ref{eq:mixed} and Eq.~\ref{eq:1} shows a tradeoff between $\mathcal{C}_{\text{large}}$ and $\mathcal{C}_{\text{few}}$.

% REMOVE ALL 'FEATURE EMBEDDING' FROM THE PAPER!!!

% \vspace{-0.05in}
\paragraph{Oracles}
% \vspace{-0.05in}
Here we explore the upper bound performance on $\mathcal{C}_{\text{few}}$.
In this setting we have all the training data for $\mathcal{C}_{\text{large}}$ and $\mathcal{C}_{\text{few}}$ in ImageNet.
For the fixed feature extractor $\mathbf{a}(\cdot)$ pre-trained on $\mathcal{D}_{\text{large}}$, we can train a linear classifier on $\mathcal{C}_{\text{large}}$ and $\mathcal{C}_{\text{few}}$, or use nearest neighbor, to see what are the upper bounds of the pre-trained $\mathbf{a}(\cdot)$. Table~\ref{tab:oracle} shows the results.
The performances are evaluated on the validation set of ILSVRC 2015~\cite{ILSVRC15} which has $50$ images for each category. The feature extractor pre-trained on $\mathcal{D}_{\text{large}}$ demonstrates reasonable accuracies on $\mathcal{C}_{\text{few}}$ which it has never seen during training for both parametric and non-parametric methods.

\begin{table}
    \small
    \centering
    \vspace{0.05in}
    \setlength{\tabcolsep}{0.5em}
    \begin{tabular}{lcccc}
        \toprule
        Classifier & Top-1 $\mathcal{C}_{\text{large}}$ & Top-5 $\mathcal{C}_{\text{large}}$ & Top-1 $\mathcal{C}_{\text{few}}$ & Top-5 $\mathcal{C}_{\text{few}}$\\
        \midrule
        NN & 70.25\% & 89.98\% & 52.46\% & 80.94 \\
        Linear & 75.20\% & 92.38\% & 60.50\% & 87.58 \\
        \bottomrule
    \end{tabular}
    \caption{Oracle 1000-way accuracies of the feature extractor $\mathbf{a}(\cdot)$ pre-trained on $\mathcal{D}_{\text{large}}$.}
    \label{tab:oracle}
\end{table}

% \vspace{-0.05in}
\paragraph{Efficiency Analysis}
% \vspace{-0.05in}
We briefly discuss the efficiencies of each method including ours on the adaptation to novel categories and the image inference.
The methods are tested on NVIDIA Tesla K40M GPUs.
For adapting to novel categories, fine-tuned ResNet and Learning like a Child require re-training the neural networks.
For re-training one epoch of the data, fine-tuned ResNet and Learning like a Child both take about 1.8 hours on 4 GPUs.
Our method only needs to predict the parameters for the novel categories using $\phi$ and add them to the original neural network.
This process takes 0.683s using one GPU for adapting the network to $100$ novel categories with one example each.
Siamese-Triplet Network and nearest neighbor with cosine distance require no operations for adapting to novel categories as they are ready for feature extraction.

For image inference, Siamese-Triplet Network and nearest neighbor are very slow since they will look over the entire dataset. Without any optimization, this can take 2.3 hours per image when we use the entire $D_{\text{large}}$.
To speed up this process in order to do comparison with ours, we first pre-compute all the features.
Then, we use a deep learning framework to accelerate the cosine similarity computation.
At the cost of 45GB memory usage and the time for feature pre-computation, we manage to lower the inference time of them to 37.867ms per image.
Fine-tuned ResNet, Learning like a Child and our method are very fast since at the inference stage, these three methods are just normal deep neural networks.
The inference speed of these methods is about 6.83ms per image on one GPU when the batch size is set to 32.
In a word, compared with other methods, our method is fast and efficient in both the novel category adaptation and the image inference.

% \vspace{-0.05in}
\paragraph{Comparing Activation Impacts}\label{sec:cai}
% \vspace{-0.05in}
In this subsection we investigate what $\phi^1$ has learned that helps it perform better than the cosine distance, which is a special solution for one-layer $\phi$ by setting $\phi$ to the identity matrix $\mathds{1}$.
\begin{figure}
    \begin{subfigure}{.455\linewidth}
        \hspace{-0.2in}\vspace{0.1in}
        \includegraphics[width=\linewidth]{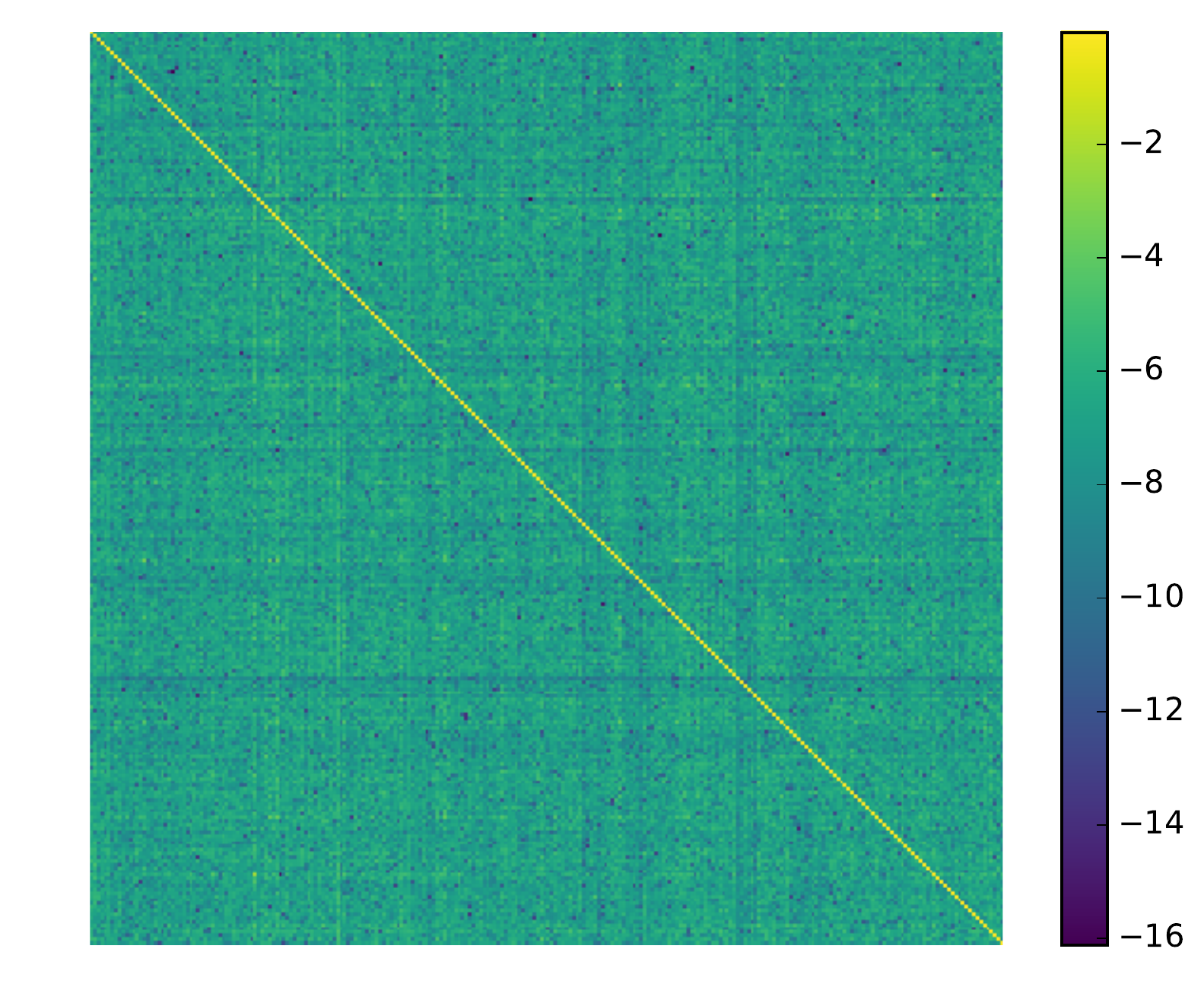}
    \end{subfigure}
    \vrule
    \begin{subfigure}{.53\linewidth}
        \hspace{0.05in}
        \includegraphics[width=\linewidth]{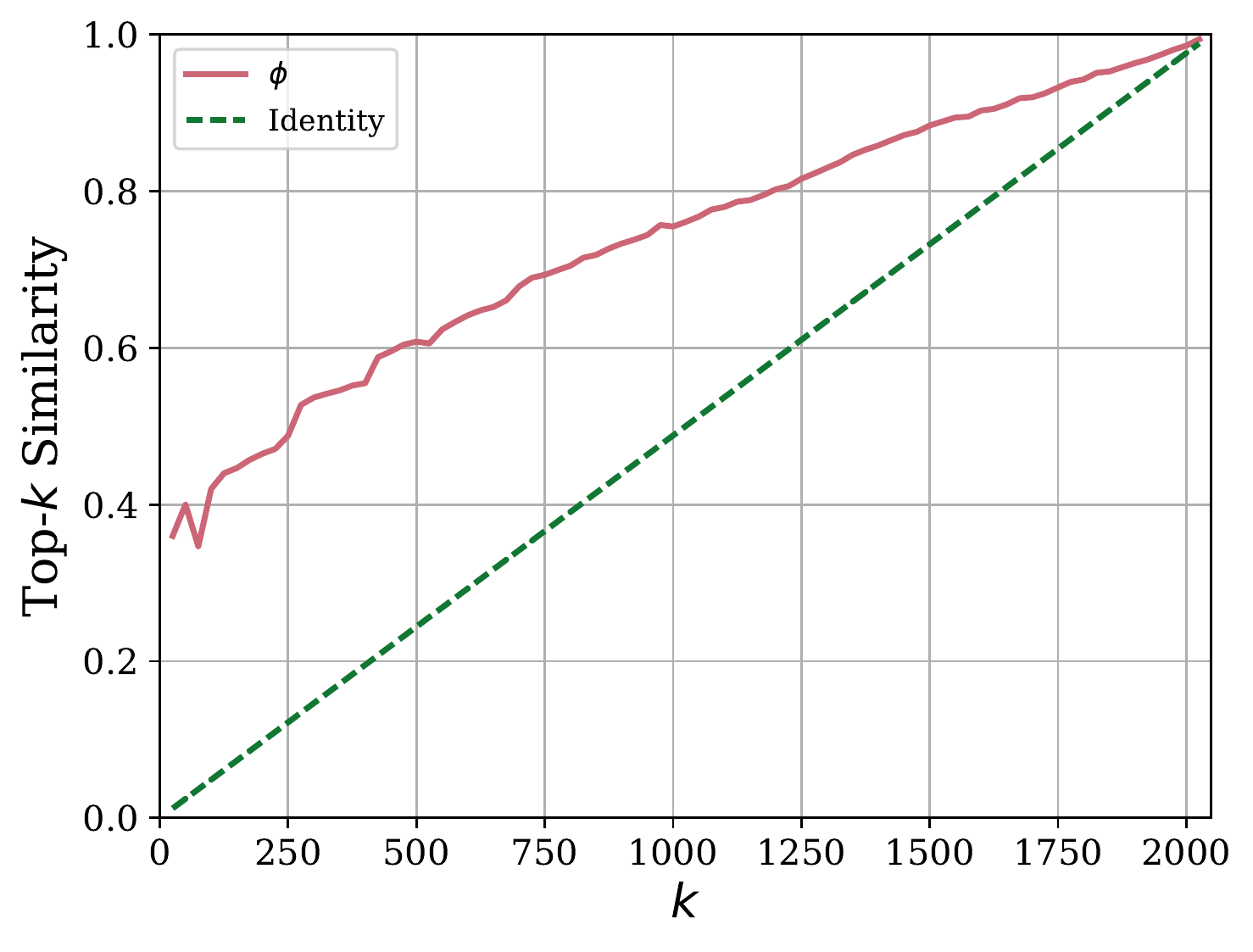}
    \end{subfigure}
    \caption{Visualization of the upper-left $256\times256$ submatrix of $\phi^1$ in $\log$ scale (left) and top-$k$ similarity between $\phi^1$, $\mathds{1}$ and \smash{$\mathbf{w}_{\text{large}}^{\text{pt}}$} (right). In the right plotting, red and solid lines are similarities between $\phi^1$ and \smash{$\mathbf{w}_{\text{large}}^{\text{pt}}$}, and green and dashed lines are between $\mathds{1}$ and \smash{$\mathbf{w}_{\text{large}}^{\text{pt}}$.}}
    \label{fig:vis}
\end{figure}
We first visualize the matrix $\phi^1_{ij}$ in $\log$ scale as shown in the left image of Figure~\ref{fig:vis}. Due to the space limit, we only show the upper-left $256\times256$ submatrix. Not surprisingly, the values on the diagonal dominates the matrix.
% We compute $|\phi^1_{ii}| / \sum_j |\phi^1_{ij}|$ for $i=1,...,2048$, and find that the mean is $0.275$ and the standard derivation is $0.067$; therefore, mathematically $\phi^1$ is not a diagonally dominant matrix. We also observe horizontal and vertical stripes in the visualization, indicating different correlations within $\mathbf{a}$.
We observe that along the diagonal, the maximum is $0.976$ and the minimum is $0.744$, suggesting that different from $\mathds{1}$, $\phi^1$ does not use each activation channel equally.
We speculate that this is because the pre-trained activation channels have different distributions of magnitudes and different correlations with the classification task.
These factors can be learned by the last fully connected layer of \smash{$\mathcal{R}_{\text{large}}^{\text{pt}}$} with large amounts of data but are assumed equal for every channel in cosine distance.
This motivates us to investigate the impact of each channel of the activation space.

For a fixed activation space, we define the \emph{impact} of its $j$-th channel on mapping $\phi$ by
$I_j(\phi)=\sum_i |\phi_{ij}|$.
Similarly, we define the activation impact $I_j(\cdot)$ on \smash{$\mathbf{w}_{\text{large}}^{\text{pt}}$} which is the parameter matrix of the last fully connected layer of \smash{$\mathcal{R}_{\text{large}}^{\text{pt}}$}. For cosine distance, $I_j(\mathds{1})=1$, $\forall j$. Intuitively, we are evaluating the impact of each channel of $\mathbf{a}$ on the output by adding all the weights connected to it.
% Since $\mathbf{w}_\text{large,pt}$ is trained for the classification task using large-amounts of data, it is reasonable to assume that $I(\mathbf{w}_\text{large,pt})$ has shown good impacts of the activation channels.
For \smash{$\mathbf{w}_{\text{large}}^{\text{pt}}$} which is trained for the classification task using large-amounts of data, if we normalize \smash{$I(\mathbf{w}_{\text{large}}^{\text{pt}})$} to unity,
the mean of \smash{$I(\mathbf{w}_{\text{large}}^{\text{pt}})$} over all channel $j$ is $2.13\text{e-}2$ and the standard deviation is $5.83\text{e-}3$.
% with the maximum of $6.80\text{e-}2$ and the minimum of $1.89\text{e-}3$.
\smash{$\mathbf{w}_{\text{large}}^{\text{pt}}$} does not use channels equally, either.

% Since $\mathbf{w}$ is well trained on the feature embedding of $\mathcal{R}_\text{large,oracle}$, we show the difference of $\phi$ and $\mathds{1}$ by comparing them with $\mathbf{w}$.
% First, we compute the cosine similarity between $I(\mathbf{a}, \phi)$, $I(\mathbf{a}, \mathbf{w})$ and $I(\mathbf{a}, \mathds{1})$.
% The cosine similarity is $0.981$ between $I(\mathbf{a}, \phi)$ and $I(\mathbf{a}, \mathbf{w})$, and $0.964$ between $I(\mathbf{a}, \mathds{1})$ and $I(\mathbf{a}, \mathbf{w})$. Therefore, $\phi$ is more similar to $\mathbf{w}$ than $\mathds{1}$. However, the difference is small.
In fact, $\phi^1$ has a high similarity with \smash{$\mathbf{w}_{\text{large}}^{\text{pt}}$}. We show this by comparing the orders of the channels sorted by their impacts.
Let $\text{top-}k(S)$ find the indexes of the top-$k$ elements of $S$. We define the top-$k$ similarity of $I(\phi)$ and \smash{$I(\mathbf{w}_{\text{large}}^{\text{pt}})$} by
\begin{equation}
  \footnotesize
  \text{OS}(\phi, \mathbf{w}_{\text{large}}^{\text{pt}}, k)=\mathbf{card}\left(\text{top-}k( I(\phi))\cap\text{top-}k(I(\mathbf{w}_{\text{large}}^{\text{pt}}))\right) / k
\end{equation}
where $\mathbf{card}$ is the cardinality of the set.
% Since $I(\mathds{1})$ does not have any informative order as they are all equal to $1$, we compute the top-$k$ similarity for it with respect to \smash{$\mathbf{w}_{\text{large}}^{\text{pt}}$} as the expected top-$k$ similarity between random order and \smash{$\mathbf{w}_{\text{large}}^{\text{pt}}$}.
The right image of Figure~\ref{fig:vis} plots the two similarities, from which we observe high similarity between $\phi$ and \smash{$\mathbf{w}_{\text{large}}^{\text{pt}}$} compared to the random order of $\mathds{1}$.
From this point of view, $\phi^1$ outperforms the cosine distance due to its better usage of the activations.

\subsection{MiniImageNet Classification}
In this subsection we compare our method with the previous state-of-the-arts on the MiniImageNet dataset.
Unlike ImageNet classification, the task of MiniImageNet is to find the correct category from $5$ candidates, each of which has $1$ example or $5$ examples for reference. The methods are only evaluated on $D_{\text{few}}$, which has $20$ categories.
For each task, we uniformly sample $5$ categories from $D_{\text{few}}$.
For each of the category, we randomly select one or five images as the references, depending on the settings, then regard the rest images of the $5$ categories as the test images.
For each task, we will have an average accuracy over this $5$ categories.
We repeat the task with different categories and report the mean of the accuracies with the $95\%$ confidence interval.

\begin{table}
    \small
    \centering
    \vspace{0.05in}
    \setlength{\tabcolsep}{0.5em}
    \begin{tabular}{lcc}
        \toprule
        Method & 1-Shot & 5-Shot \\
        \midrule
        Fine-Tuned Baseline & 28.86 $\pm$ 0.54\% & 49.79 $\pm$ 0.79\% \\
        Nearest Neighbor & 41.08 $\pm$ 0.70\% & 51.04 $\pm$ 0.65\% \\
        Matching Network~\cite{DBLP:conf/nips/VinyalsBLKW16} & 43.56 $\pm$ 0.84\% & 55.31 $\pm$ 0.73\% \\
        Meta-Learner LSTM~\cite{ravi2017optimization} & 43.44 $\pm$ 0.77\% & 60.60 $\pm$ 0.71\% \\
        MAML~\cite{pmlr-v70-finn17a} & 48.70 $\pm$ 1.84\% & 63.11 $\pm$ 0.92\% \\
        \hdashline
        Ours-Simple & {\bf\color{blue}54.53} $\pm$ 0.40\% & {\bf\color{blue}67.87} $\pm$ 0.20\% \\
        Ours-WRN & {\bf\color{red}59.60} $\pm$ 0.41\% & {\bf\color{red}73.74} $\pm$ 0.19\% \\
        \bottomrule
    \end{tabular}
    \caption{$5$-way accuracies on MiniImageNet with $95\%$ confidence interval. Red: the best, and blue: the second best.}
    \label{tab:mini}
\end{table}

Table~\ref{tab:mini} summarizes the few-shot accuracies of our method and the previous state-of-the-arts.
For fair comparisons, we implement two convolutional neural networks.
The convolutional network of Ours-Simple is the same as that of Matching Network~\cite{DBLP:conf/nips/VinyalsBLKW16} while Ours-WRN uses WRN-28-10~\cite{Zagoruyko2016WRN} as stated in \S\ref{sec:id}.
The experimental results demonstrate that our average accuracies are better than the previous state-of-the-arts by a large margin for both the Simple and WRN implementations.

It is worth noting that the methods \cite{pmlr-v70-finn17a,ravi2017optimization,DBLP:conf/nips/VinyalsBLKW16} are not evaluated in the full ImageNet classification task.
This is because the architectures of these methods, following the problem formulation of Matching Network~\cite{DBLP:conf/nips/VinyalsBLKW16}, can only deal with the test tasks that are of the same number of reference categories and images as that of the training tasks, limiting their flexibilities for classification tasks of arbitrary number of categories and reference images.
In contrast, our proposed method has no assumptions regarding the number of the reference categories and the images, while achieving good results on both tasks.
From this perspective, our methods are better than the previous state-of-the-arts in terms of both the performance and the flexibility.

% \vspace{-0.05in}
\section{Conclusion}\label{sec:c}
% \vspace{-0.05in}
In this paper, we study a novel problem: can we develop a unified approach that works for both large-scale and few-shot learning.
Our motivation is based on the observation that in the final classification layer of a pre-trained neural network, the parameter vector and the activation vector have highly similar structures in space.
This motivates us to learn a category-agnostic mapping from activations to parameters.
Once this mapping is learned, the parameters for any novel category can be predicted by a simple forward pass, which is significantly more convenient than re-training used in parametric methods or enumeration of training set used in non-parametric approaches.

We experiment our novel approach on the MiniImageNet dataset and the challenging full ImageNet dataset. The challenges of the few-shot learning on the full ImageNet dataset are from the large number of categories (1000) and the very limited number ($<4$) of training samples for $\mathcal{C}_{\text{few}}$.
On the full ImageNet dataset, we show promising results, achieving state-of-the-art classification accuracy on novel categories by a significant margin while maintaining comparable performance on the large-scale classes.
We further visualize and analyze the learned parameter predictor, as well as demonstrate the similarity between the predicted parameters and those of the classification layer in the pre-trained deep neural network in terms of the activation impact.
On the small MiniImageNet dataset, we also outperform the previous state-of-the-art methods by a large margin.
The experimental results demonstrate the effectiveness of the proposed method for learning a category-agnostic mapping.

% In the future we plan to extend this framework to zero-shot learning, where the parameters are initialized as vectors in a different modality.

{\small
\bibliographystyle{ieee}
\bibliography{egbib}
}

\end{document}